\documentclass{article} 
\usepackage{arxiv}
 \usepackage{graphicx}

\usepackage{booktabs}
\usepackage{url}
\usepackage{bookmark}
\usepackage{color}

\usepackage{hyperref}
\date{}
\title{Forecasting Open-Weight AI Model Growth on HuggingFace}

      \author{Kushal Raj Bhandari\\
        Department of Computer Science\\
        Network Science and Technology Center\\
        Rensselaer Polytechnic Institute\\
        Troy, NY, USA \\
        \texttt{bhandk@rpi.edu}\\
        \And
        Pin-Yu Chen\\
        IBM Research \\
        Yorktown Heights, NY, USA \\
        \texttt{pin-yu.chen@ibm.com}\\
        \And 
         Jianxi Gao\\
        Department of Computer Science\\
        Network Science and Technology Center\\
        Rensselaer Polytechnic Institute\\
        Troy, NY, USA \\
        \texttt{gaoj8@rpi.edu}\\
        }

\begin{document}

    \maketitle

    \begin{abstract}
    As the open-weight AI landscape continues to proliferate—with model development, significant investment, and user interest—it becomes increasingly important to predict which models will ultimately drive innovation and shape AI ecosystems. Building on parallels with citation dynamics\cite{wangQuantifyingLongTermScientific2013} in scientific literature, we propose a framework to quantify how an open-weight model’s influence evolves. Specifically, we adapt the model introduced by Wang et al. for scientific citations, using three key parameters—immediacy, longevity, and relative fitness—to track the cumulative number of fine-tuned models of an open-weight model. Our findings reveal that this citation-style approach can effectively capture the diverse trajectories of open-weight model adoption, with most models fitting well and outliers indicating unique patterns or abrupt jumps in usage. Link to the website for trajectory visualization: 
    \footnote{\url{https://forecasthuggingfacemodels.onrender.com/}}
    \end{abstract}

    \section{Introduction}
       The rapid expansion of the open-weight AI ecosystem has led to a diverse landscape of models, each varying in size, company affiliation, and adoption patterns, raising critical questions about their long-term influence and impact \cite{bommasaniOpportunitiesRisksFoundation2022}. Understanding how influential a model will become is crucial for AI governance, business strategy, and scientific progress \cite{costaDemocratizationArtificialIntelligence2024, luitseGreatTransformerExamining2021, euArtificialIntelligenceAct, maslejArtificialIntelligenceIndex2024}. 
        \begin{figure}[ht]
            \begin{center}
                \includegraphics[width=\linewidth]{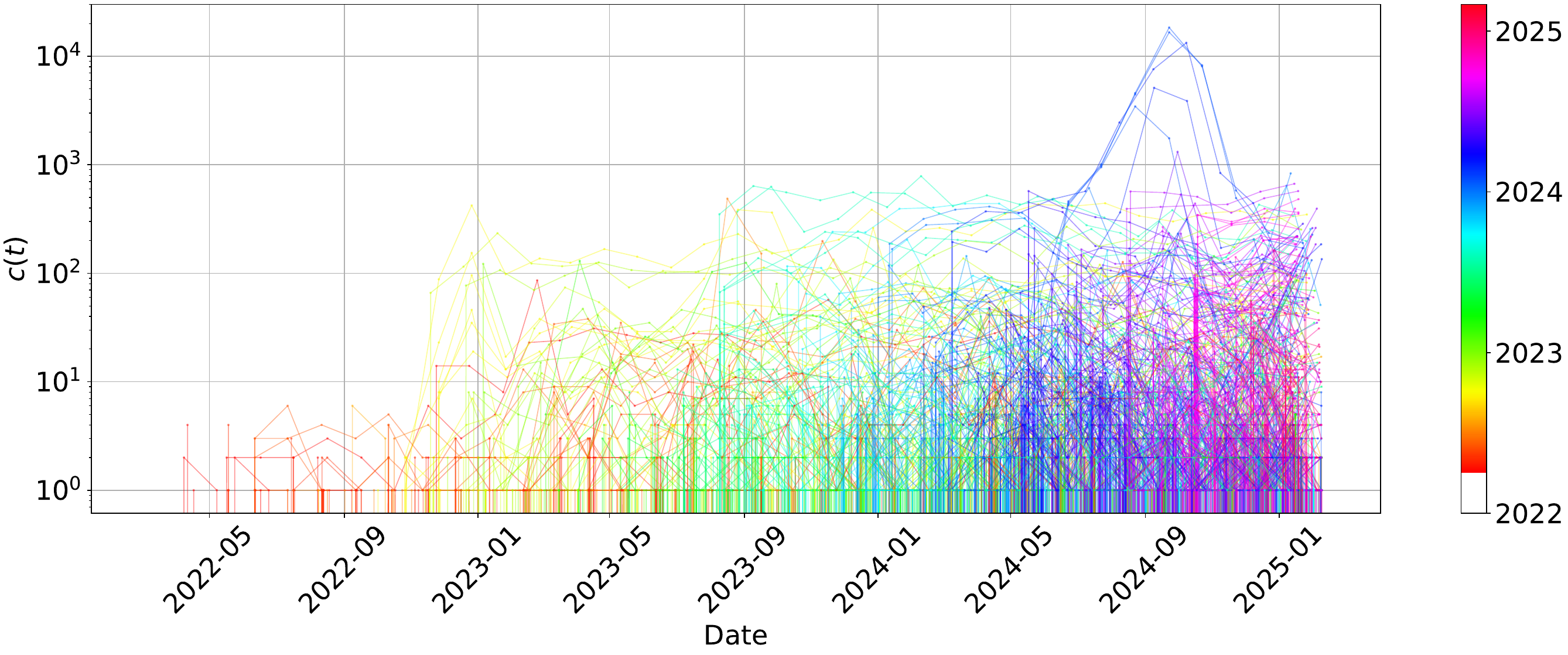}
            \end{center}
            \caption{Monthly number of fine-tuned models after a base model’s release, with colors denoting the time when it was created. }
            \label{fig:monthly_finetune_number}
        \end{figure}
        
        Recent work underscores the growing impact of open-weight foundation models, highlighting both their benefits and challenges. Henderson et al. \cite{hendersonFoundationModelsFair2023} examine fair use uncertainties when training on copyrighted material, while Chan et al. \cite{chanHazardsIncreasinglyAccessible2023} address both the perils of fine-tuning accessible models and the importance of governance mechanisms, as further emphasized by Chan et al. \cite{chanVisibilityAIAgents2024} in their discussion of agent identifiers and real-time monitoring. Eiras \cite{eirasRisksOpportunitiesOpenSource2024} evaluates risks and opportunities across various development stages, advocating open-sharing practices alongside mitigation strategies. Collectively, these works illustrate the need for thoughtful consideration of open-source AI’s evolving role in research, industry, security, and governance. 
       
       A closer look at fine-tuning patterns (figure~\ref{fig:monthly_finetune_number}) highlights the diversity of open-weight model adoption. Some base models experience rapid adoption, while others grow steadily. Understanding these dynamics is key to anticipating model prominence. Motivated by these insights, \textbf{can we predict the trajectory of influence an open-weight model will have on the AI community?} This inquiry drives our exploration into utilizing early adoption trends—particularly the observed growth rates—which can reliably forecast long-term impact, ultimately informing both strategic decisions and governance in the AI domain.

    \section{Framework for Analysis} \label{sec:framework}
     
       Building on this quantitative perspective, parallels emerge with the citation dynamics observed in scientific research\cite{wangQuantifyingLongTermScientific2013}. We utilize the dynamics of the citation model proposed by Wang et al., hypothesizing that a similar approach can effectively capture open-weight model adoption due to analogous patterns observed in usage. Specifically, we propose a framework defined by three key parameters—immediacy ($\mu_i$), longevity ($\sigma_i$), and relative fitness ($\lambda_i$)—along with $t$, the time duration after a model is released, $m$, the average number of fine-tuned models, and $\Phi$, the cumulative normal distribution function. The adoption dynamics of an open-weight model are governed by
        \begin{equation} 
            c_i^t = m \left(e^{\lambda_i \Phi \left( \frac{\ln t - \mu_i}{\sigma_i} \right) } - 1 \right), 
            \label{eq:citation_model}
        \end{equation}
        and, 
        \begin{equation}
            \Phi(x) = \frac{1}{\sqrt{2\pi}} \int_{-\infty}^{x} \exp\!\left(-\frac{t^2}{2}\right) dt,
            \label{eq:cumm_normal_distribution}
        \end{equation}
        where immediacy ($\mu_i$) governs the time required for an open-weight model to reach its peak adoption, longevity ($\sigma_i$) captures the decay rate of the model’s influence, and relative fitness ($\lambda_i)$ represents the model’s inherent relative influence compared to other models in the ecosystem.

        This dynamics effectively captures how attention and usage evolve in scenarios where an innovation—a scientific publication or an open-weight model—is introduced and disseminated across a community. Similar to how a citation grows rapidly once a paper is published and then gradually decays over time, open-weight models also tend to experience an initial burst of interest followed by sustained but diminishing engagement. By incorporating parameters like immediacy, longevity, and relative fitness, the model reflects both the short-term surge in popularity and the long-term retention trajectory. Consequently, this aligns well with observed patterns in open-source development communities, making it a suitable framework for understanding and predicting model usage. Figure \ref{fig:monthly_cummulative_finetune_number}, similar to figure \ref{fig:monthly_finetune_number}, illustrates the actual cumulative number of fine-tuned models after the base model is open source through HuggingFace. 
        
        Following a similar approach to modeling the cumulative number of fine-tuned models, we also examine the cumulative download trajectories for each model. Using the framework, in Appendix Section \ref{sec:downloads}, we empirically predict the cumulative number of downloads for the widely popular DeepSeek models. Given 20 days of downloads for \texttt{deepseek-ai/DeepSeek-R1}, we expect it to reach 1.3 billion cumulative downloads within 75 days.

        \subsection{Fitting the model to Empirical Data}

        Using the HuggingFace ecosystem\footnote{\url{https://huggingface.co/}}, we fit equation \ref{eq:citation_model} to track the cumulative number of fine-tuned models built from a base model $i$, measured $t$ months after its release as an open-weight model on HuggingFace (Detailed in Appendix \ref{sec:data_collection}). 
        
        \begin{figure}[ht]
            \begin{center}
                \includegraphics[width=\linewidth]{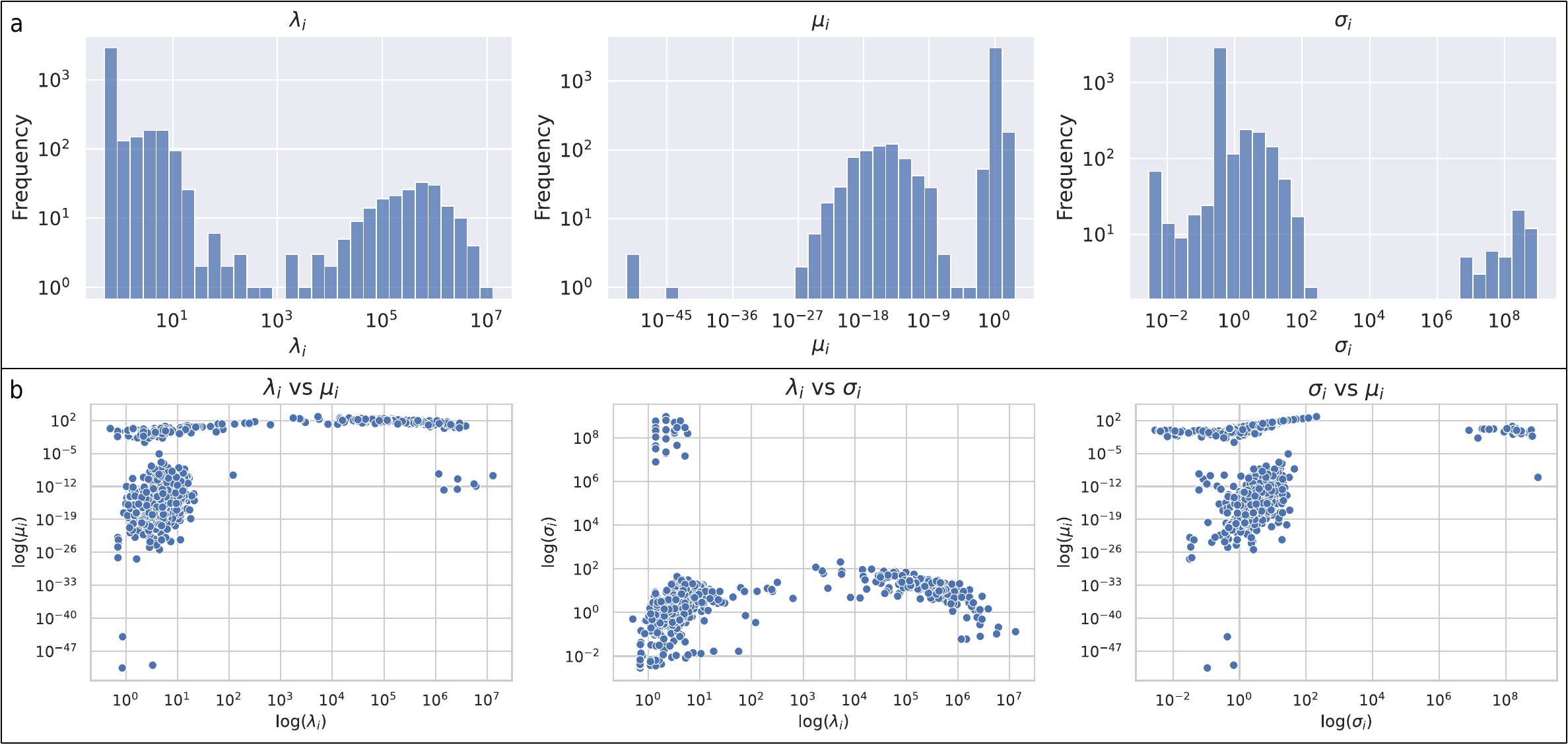}
            \end{center}
            \caption{(\textbf{a}) Distribution of values for $\lambda$, $\mu$, and $\sigma$. (\textbf{b}) Pairwise relationships among immediacy ($\mu_i$), longevity ($\sigma_i$), and relative fitness ($\lambda_i$) on log-scale axes.}
            \label{fig:parameter}
        \end{figure}
        From figure \ref{fig:parameter}a, we can see that most models cluster around relatively narrow bands of $\lambda$, $\mu$, and $\sigma$, yet there are notable outliers where the parameter estimates are extremely large or small. In the $\lambda_i$ histogram, many models have values concentrated near the lower end (close to 1 or below), but several parameter fits extend out to $10^5$ or $10^7$. Similarly, $\mu_i$ is heavily concentrated in a narrow region around $10^{-4}$, with a few models at exceedingly small values (e.g., $10^{-14}$ or $10^{-44}$), while $\sigma_i$ exhibits a main peak around $0.1$ and occasional jumps to values near $10^2$ or $10^8$. 

        This pattern indicates that the citation‐style adoption curve fits reasonably well for most models, yielding parameter values in plausible ranges that reflect gradual growth, moderate time shifts, and a smooth transition to saturation. However, for a minority of models, the curve either overcompensates or cannot accurately capture the sudden jumps or unusual trajectories, forcing the optimizer to return extreme parameter estimates. These outliers underscore that while the model works broadly well, certain edge cases or atypical usage patterns may require additional constraints or alternative modeling assumptions to avoid unrealistic parameter values.
    \subsection{Dependency of parameters}
        While each parameter provides a distinct lens into model adoption—$\mu_i$ dictating time to peak, $\sigma_i$ controlling longevity, and $\lambda_i$ determining overall attractiveness—their relationships reveal more nuanced dynamics. For instance, models with high $\lambda_i$ but low $\sigma_i$ experience rapid adoption but short-lived influence, suggesting strong initial appeal but limited long-term utility. Conversely, models with moderate $\lambda_i$ and high $\sigma_i$ exhibit sustained adoption, indicating persistent engagement over time despite lacking an immediate surge in popularity. These dependencies illustrate how different models follow varied life cycles, from fleeting trends to enduring contributions, emphasizing the importance of modeling adoption beyond single-parameter distributions.
        
        Figure \ref{fig:parameter}b shows the pairwise relationships among the three parameters—immediacy $\mu_i$, longevity $\sigma_i$, and relative fitness $\lambda_i$—plotted on log‐scale axes. In the left panel ($\lambda_i$ vs. $\mu_i$), we see that some models combine very high relative fitness with smaller values of $\mu_i$, indicating both a rapid rise to peak adoption and strong overall influence. Others stretch to larger $\mu_i$, implying they take longer to reach peak usage even if they remain moderately or highly appealing. In the middle panel ($\lambda_i$ vs. $\sigma_i$), points with high $\lambda_i$ but small $\sigma_i$ suggest that although the model attracts many users, its influence weakens more quickly, whereas higher $\sigma_i$ corresponds to a more prolonged “tail” of adoption. The right panel ($\sigma_i$ vs. $\mu_i$) highlights that even models with similar times to peak can exhibit widely different decay rates—some fade rapidly while others remain in use for much longer. The broad range of parameter values across multiple orders of magnitude demonstrates the diversity of open‐source adoption trajectories and highlights the framework's flexibility in capturing such heterogeneous dynamics.
        
    \section{Organization-Specific analysis on model's importance relative to other models}
        The diversity of parameter distributions in figure \ref{fig:parameter}(\textbf{a}) underscores the varied pathways through which open-weight models gain and sustain adoption. While some models achieve rapid prominence with enduring influence, others experience a slower ascent or a more transient impact. These heterogeneous adoption dynamics suggest that intrinsic qualities do not solely dictate model success but are also shaped by external factors, including company strategies and ecosystem positioning.
        \begin{figure}[h]
            \begin{center}
                \includegraphics[width=\linewidth]{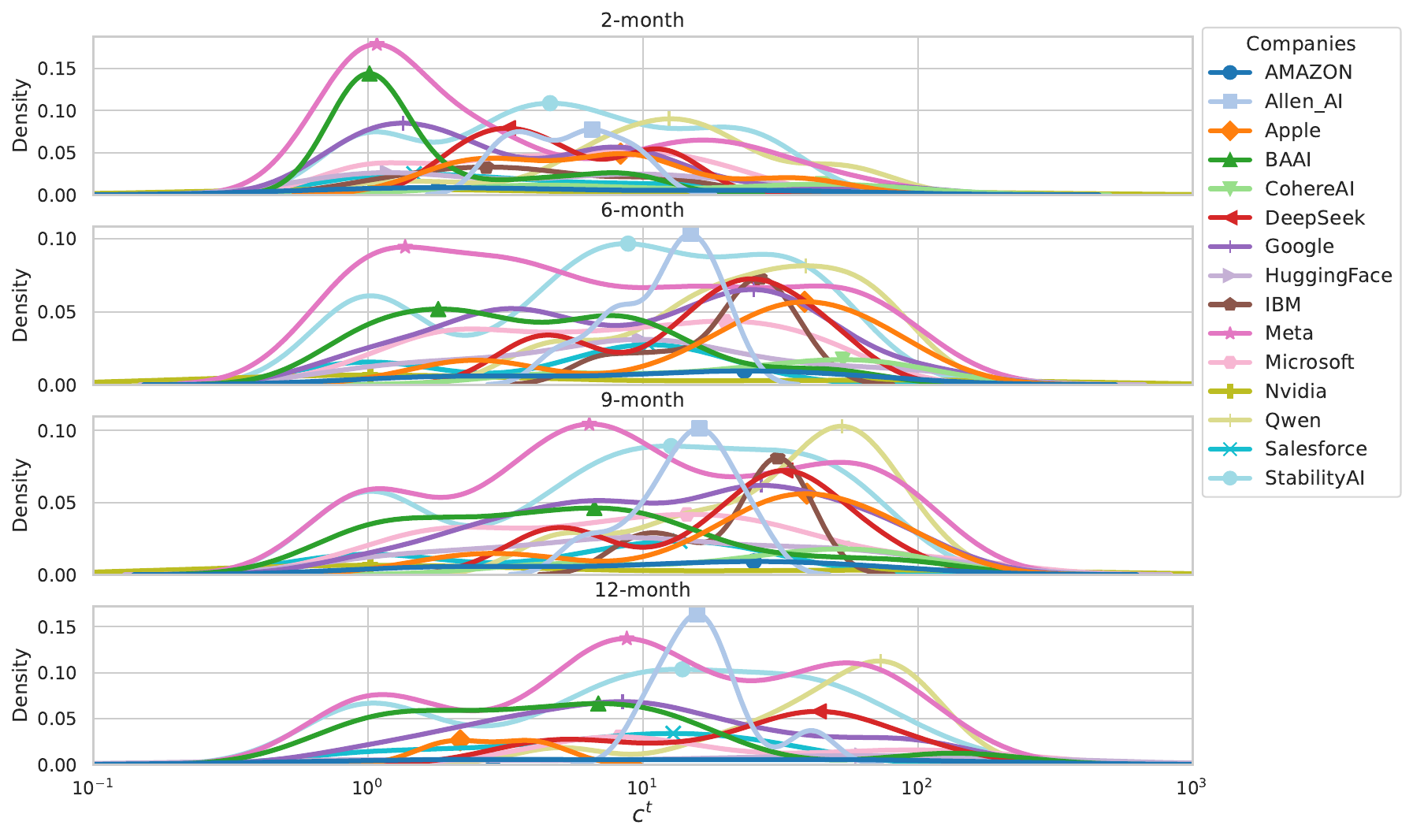}
            \end{center}
            \caption{Density plots illustrating the cumulative number of fine-tuned models for relative fitness of ($1 \leq \lambda_i \leq 10$) at the 2-month, 6-month, and 12-month marks, segmented by companies. IBM and CohereAI are omitted from the 12-month plot due to the absence of models older than 12 months.}
            \label{fig:distribution_time_period}
        \end{figure}
        
         By examining models with moderate relative fitness ($1 \leq \lambda_i \leq 10$), figure \ref{fig:distribution_time_period} provides insights into the temporal shifts in the frequency of fine-tuned models, revealing how different organizations’ models evolve in their attractiveness for general adaptation over time. At the early stage (2 months), distinct peaks in $c^t$ emerge, suggesting that models in the fitness range receive some degree of initial interest for some organizations and some have some more outstanding interests. Over time, the KDE curves for most companies begin to align in broader but partially overlapping bands, indicating that short‐term disparities in how quickly each base model is fine‐tuned start to level out. 
         Meta, BAAI, and StabilityAI exhibit the highest peaks at low $c^t$, indicating that a significant number of base models are released by these companies. At the same time, companies like StabilityAI and Qwen display broader distributions, suggesting variability in adoption, and Microsoft, Amazon, and Apple have minimal early fine-tuning adoption.
        
        The gradual consolidation, where peaks shift toward higher $c^t$ values for some models and compress for others, reflects both sustained adoption and potential “latecomer” effects. This behavior is consistent with the idea that models with $\lambda_i$ in the relatively higher range dominate immediately; rather, they accumulate a steady stream of fine‐tuned variants as the model gets widely adopted.  At 6 months and 9 months, Meta, BAAI, and StabilityAI still show strong peaks at lower $c^t$, but Allen\_AI and Qwen exhibit sharper peaks at higher $c^t$, suggesting that some of their base models gain significant traction. We also see the ``latecomer" effect with companies like IBM, which had little to no adoption during earlier days but has significant interest after certain months. By 12 months, the distributions consolidate, indicating more widespread fine-tuning across different base models, with companies like Meta, StabilityAI, and Qwen showing extended tails, signifying that some models continue accumulating fine-tuned variants over time.
        
        These density curves reinforce the notion that models with high relative fitness levels follow the number of fine-tuning trajectories that differ initially but increasingly resemble one another over longer horizons. This aligns with expectations from the citation model \cite{wangQuantifyingLongTermScientific2013}, where cumulative citations of papers with early “head starts” or slower uptake subside, comparable fitness values dominate the eventual adoption outcome, yielding nearly the same ultimate impact $c^t$. Individual trajectories of models based on these companies are highlighted within Appendix \ref{sec:individual_cummulative_citation}.
    \section{Conclusion}
        In this study, we adapted a well-known citation model to examine the adoption dynamics for open-source models. Through this analogy, we introduced three parameters—immediacy, longevity, and relative fitness—to characterize early and sustained usage patterns. Our empirical analysis of monthly fine-tuning counts reveals that most models exhibit predictable trajectories, whereas a minority of AI models show sudden, significant surges in popularity, challenging the assumption of simple exponential decay. We also found that organization-level factors significantly shape these usage patterns: models released by Meta, Google, BAAI, and StabilityAI display distinct adoption curves reflective of each company’s open-source strategies and ecosystem support. Our citation-inspired framework helps stakeholders—from industry leaders optimizing release strategies to policymakers overseeing AI governance—better understand how open-weight models evolve and gain traction. Future research can extend this approach by integrating additional data to refine long-term adoption forecasts and further clarify the influences driving model success.
        

\clearpage

\appendix
\section*{Appendix}
\section{Data Collection}\label{sec:data_collection}
 \begin{figure}[ht]
        \begin{center}
            \includegraphics[width=\linewidth]{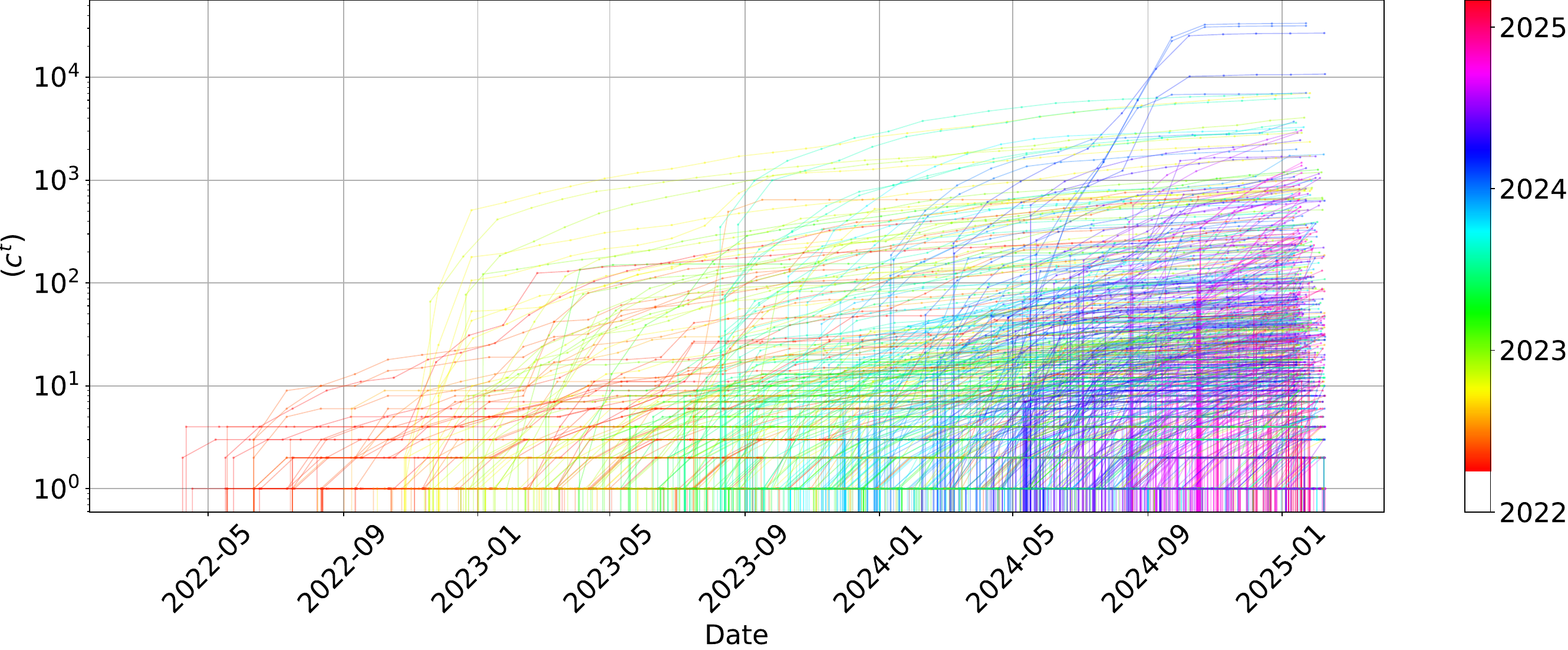}
        \end{center}
        \caption{Monthly cumulative number of fine-tuned models following the release of the base model, with colors indicating the base models' creation years, illustrating trends in fine-tuning patterns over time.}
        \label{fig:monthly_cummulative_finetune_number}
    \end{figure}

We collect data on open-weight model adoption using the HuggingFace API, which provides comprehensive metadata on models uploaded to the platform. Given that HuggingFace\footnote{https://huggingface.co/} serves as the primary repository for open-source AI models, we assume that the vast majority of publicly available models are hosted there .

To quantify fine-tuning activity, we track the number of fine-tuned models uploaded to HuggingFace after the release of a given base model, aggregating counts by month. For example, since LLaMA 2 was released on July 18, 2023, we begin counting fine-tuned models from that point onward. However, we exclude the earliest models such as GPT-2 and BERT variants, as these were only uploaded to HuggingFace on March 2, 2022, despite being released much earlier, which could distort the adoption timeline.

Fine-tuned models are identified based on their tags and model names, but due to inconsistencies in labeling, it is possible that some fine-tuned models are not captured in our dataset. Additionally, HuggingFace provides only the total number of downloads for a model without historical breakdowns. To address this, we began recording daily total downloads for each model starting in September 2024, allowing us to approximate temporal trends in adoption.

\begin{figure}[ht]
        \begin{center}
            \includegraphics[width=\linewidth]{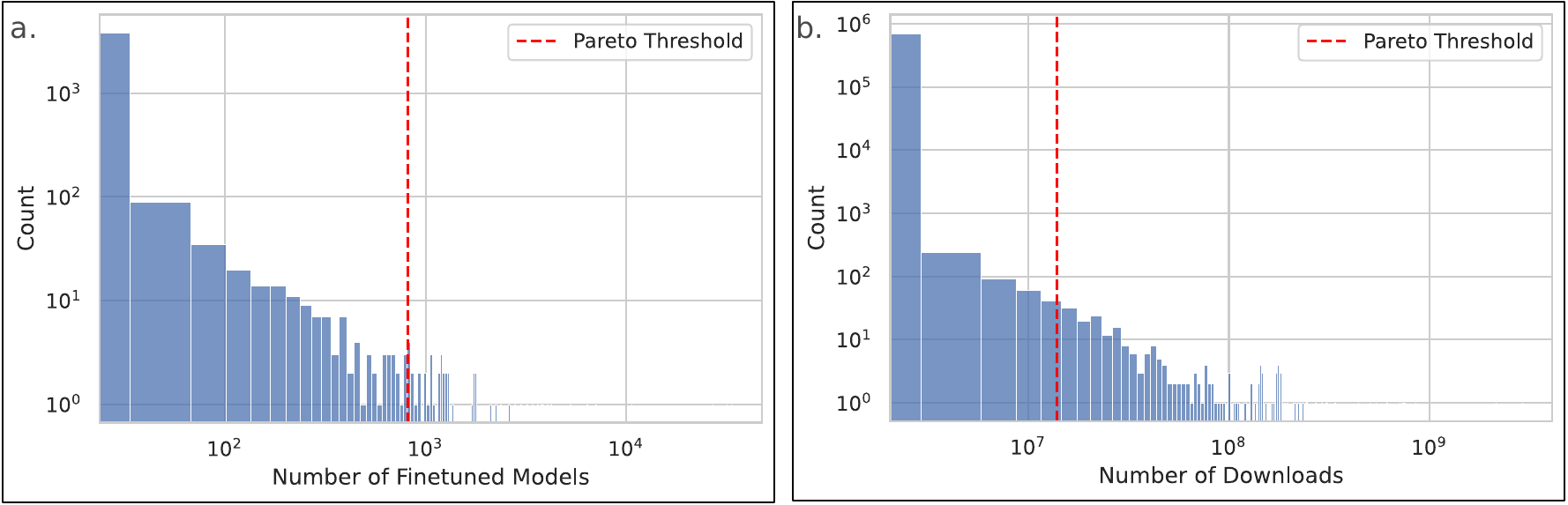}
        \end{center}
        \caption{Distribution of downloads and fine-tuned models across Hugging Face models, illustrating a strong Pareto-like concentration. A small fraction of models (1.26\% for fine-tuning, 0.035\% for downloads) accounts for 80\% of total activity, highlighting the dominance of a few models in the open-source AI ecosystem.}
        \label{fig:distribution}
    \end{figure}

Figure \ref{fig:distribution} plots the distribution of both the number of downloads and the number of fine-tuned models across different models on HuggingFace that we use for analysis, exhibits a strong skew, aligning with the well-documented Pareto principle, often referred to as the ``80-20 rule." A remarkably small fraction of models account for the vast majority of activity: only 1.26\% (52 out of 4,121) of models represent 80\% of all fine-tuned models, while an even smaller 0.035\% (249 out of 714,018) of models contribute to 80\% of all downloads. This suggests that model usage follows a highly skewed distribution, where a handful of models dominate the ecosystem while the vast majority remain relatively underutilized. 

\section{Parameter $m$ and $t$} \label{sec:parameter_m_t}
In our adaptation of the citation model, we set $ m = 1 $, meaning that the model predicts the cumulative number of fine-tuned models directly without scaling by an arbitrary reference count. This differs from the original citation model, where Wang et al. \cite{wangQuantifyingLongTermScientific2013} set $ m = 30 $ to account for the typical number of references in a new paper. They noted that fixing $ m $ for all papers allows for easy parameter comparison and that increasing $ m $ results in a smaller $\lambda$ but does not affect the overall fitting or prediction when $ m $ is comparable to the average number of citations per paper. In our case, because fine-tuning does not have a well-defined equivalent to the number of references in a paper, we normalize $ m $ to 1, making the adoption curve directly interpretable in terms of the absolute count of fine-tuned models. This ensures that $\lambda_i$ captures the relative fitness of each base model without needing an external scaling factor.

Additionally, we count $ t $ in months rather than the year used in the original citation model. This adjustment reflects the timescale of model adoption, where fine-tuned models typically emerge over weeks and months rather than daily fluctuations. By aggregating over months, we reduce noise and better capture long-term adoption trends, aligning with the natural timescale at which open-weight models gain traction within the community.

\section{Fitting Equation \ref{eq:citation_model} on empirical data}
\begin{figure}[ht]
    \begin{center}
        \includegraphics[width=\linewidth]{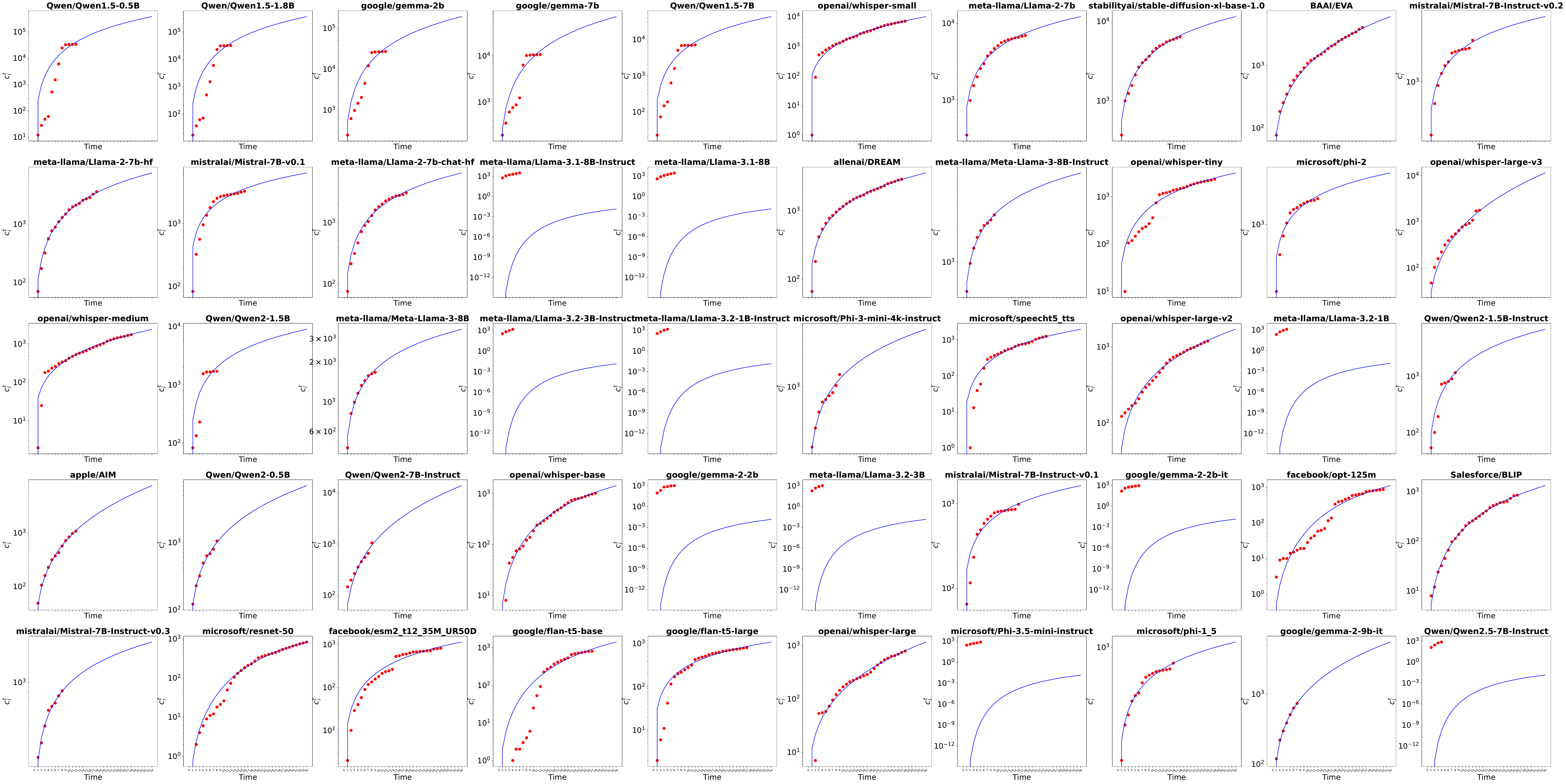}
    \end{center}
    \caption{Each subplot represents models, where the x-axis denotes the time, t(month), after release, and the y-axis represents the cumulative count ($c_i^t$) on a logarithmic scale. Red dots indicate empirical data points, while blue curves correspond to the fitted function using the extracted parameters ($\lambda_i$, $\mu_i$, $\sigma_i$).}
    \label{fig:top_50_most_finetuned}
\end{figure}
The figure illustrates the cumulative adoption trends of various AI models over time, with each subplot representing a specific model. The x-axis denotes time, while the y-axis shows the cumulative count on a logarithmic scale. Red dots indicate empirical data points, and the blue curves correspond to the fitted functions using the extracted parameters: $\lambda_i$ (growth rate), $\mu_i$ (shift parameter), and $\sigma_i$ (scaling factor). The overall fit demonstrates that the selected functional form effectively captures cumulative growth patterns for most models.

The table \ref{tab:model_parameter_summary} includes the extracted parameters for the models listed in figure \ref{fig:top_50_most_finetuned}. The parameters reveal considerable variation in the cumulative number of fine-tuned model trajectories. Models such as \texttt{openai/whisper-large-v3}, \texttt{BAAI/EVA}, and \texttt{microsoft/Phi-3-mini-4k-instruct} exhibit high $\lambda_i$ values, signifying rapid cumulative adoption. These models also have relatively high $\mu_i$ values, indicating an early adoption surge, likely due to strong initial interest and high demand. 

Notably, for some models like \texttt{meta-llama/Llama-3.1-8B} and \texttt{microsoft/Phi-3.5-mini-instruct}, highlighted by ``*'' in table \ref{tab:model_parameter_summary}, display lower $\lambda_i$ and $\sigma_i$ values, reflecting a more gradual accumulation of usage over time. The parameter estimation resulted in $\lambda_i = 0.5$, $\mu_i = 2.0$, and $\sigma_i = 0.5$. These values suggest that the equation used for fitting failed to accurately capture the underlying growth dynamics for those models, leading to an ineffective approximation of their cumulative adoption curves. This discrepancy is due to irregular growth trends, as seen in the figure \ref{fig:top_50_most_finetuned}. This suggests that the framework \ref{eq:citation_model} still requires some more comprehensive adaptation since it fails to capture such a growth trend. 

The visualization highlights diverse adoption patterns among AI models, with some experiencing rapid early adoption and subsequent saturation, while others follow a steady and prolonged growth trajectory. Deviations from the fitted curves in certain cases suggest additional influencing factors, such as accessibility, licensing constraints, or specific application domains. These findings offer valuable insights into AI model adoption, providing a quantitative framework for assessing their long-term impact and diffusion.

\begin{table}[ht]
    \centering
    \small
    \setlength{\tabcolsep}{5pt} 
    \renewcommand{\arraystretch}{1.2} 
    \scalebox{0.8}{
    \begin{tabular}{lccc}
        \toprule
        \textbf{Model Name} & $\lambda_i$ & $\mu_i$ & $\sigma_i$ \\
        \midrule
        Qwen/Qwen1.5-0.5B & 21.2340 & 1.18e-15 & 3.9044 \\
        Qwen/Qwen1.5-1.8B & 21.1198 & 1.00e-15 & 3.8795 \\
        google/gemma-2b & 20.7799 & 2.56e-14 & 4.8182 \\
        google/gemma-7b & 18.9374 & 9.78e-15 & 4.5854 \\
        Qwen/Qwen1.5-7B & 18.0948 & 1.41e-19 & 4.6136 \\
        openai/whisper-small & 294604.7393 & 90.9031 & 22.4477 \\
        meta-llama/Llama-2-7b & 17.2144 & 1.04e-17 & 8.8424 \\
        stabilityai/stable-diffusion-xl-base-1.0 & 16.9046 & 5.80e-11 & 7.8304 \\
        BAAI/EVA & 454253.6120 & 95.8721 & 23.0329 \\
        mistralai/Mistral-7B-Instruct-v0.2 & 16.1882 & 7.18e-15 & 7.7386 \\
        meta-llama/Llama-2-7b-hf & 15.3191 & 1.76e-14 & 4.9636 \\
        mistralai/Mistral-7B-v0.1 & 15.9177 & 1.03e-15 & 8.2057 \\
        meta-llama/Llama-2-7b-chat-hf & 15.2853 & 9.88e-12 & 5.5452 \\
        meta-llama/Llama-3.1-8B-Instruct & 0.5* & 2.0* & 0.5* \\
        meta-llama/Llama-3.1-8B & 0.5* & 2.0* & 0.5* \\
        allenai/DREAM & 24.2332 & 4.9102 & 9.2243 \\
        meta-llama/Meta-Llama-3-8B-Instruct & 15.9664 & 1.47e-10 & 10.6965 \\
        openai/whisper-tiny & 13.4653 & 2.04e-15 & 4.1449 \\
        microsoft/phi-2 & 15.2437 & 8.83e-18 & 9.5035 \\
        openai/whisper-large-v3 & 528070.6635 & 66.4680 & 15.8209 \\
        openai/whisper-medium & 460695.9213 & 88.9759 & 21.2067 \\
        Qwen/Qwen2-1.5B & 16.0543 & 4.44e-12 & 6.1988 \\
        meta-llama/Meta-Llama-3-8B & 15.2420 & 1.06e-10 & 11.5625 \\
        meta-llama/Llama-3.2-3B-Instruct & 0.5* & 2.0* & 0.5* \\
        meta-llama/Llama-3.2-1B-Instruct & 0.5* & 2.0* & 0.5* \\
        microsoft/Phi-3-mini-4k-instruct & 114364.7070 & 142.1125 & 37.0978 \\
        microsoft/speecht5\_tts & 12.3327 & 6.40e-10 & 3.5563 \\
        openai/whisper-large-v2 & 68.7205 & 13.4940 & 10.0765 \\
        meta-llama/Llama-3.2-1B & 0.5* & 2.0* & 0.5* \\
        Qwen/Qwen2-1.5B-Instruct & 15.1078 & 1.70e-17 & 4.9109 \\
        apple/AIM & 120131.6996 & 66.9603 & 17.3784 \\
        Qwen/Qwen2-0.5B & 32058.6364 & 76.6518 & 21.8903 \\
        Qwen/Qwen2-7B-Instruct & 415361.3050 & 78.3713 & 18.9740 \\
        openai/whisper-base & 11.2185 & 6.13e-20 & 2.7420 \\
        google/gemma-2-2b & 0.5* & 2.0* & 0.5* \\
        meta-llama/Llama-3.2-3B & 0.5* & 2.0* & 0.5* \\
        mistralai/Mistral-7B-Instruct-v0.1 & 13.4460 & 7.33e-15 & 8.2182 \\
        google/gemma-2-2b-it & 0.5* & 2.0* & 0.5* \\
        facebook/opt-125m & 9.2155 & 1.68e-14 & 1.4702 \\
        Salesforce/BLIP & 11.6421 & 0.2335 & 2.7321 \\
        mistralai/Mistral-7B-Instruct-v0.3 & 14.0439 & 3.31e-09 & 7.2751 \\
        microsoft/resnet-50 & 9.0884 & 4.48e-21 & 1.6266 \\
        facebook/esm2\_t12\_35M\_UR50D & 11.4140 & 6.74e-19 & 3.5063 \\
        google/flan-t5-base & 10.3708 & 1.28e-19 & 1.9899 \\
        google/flan-t5-large & 11.8440 & 8.27e-14 & 4.6042 \\
        openai/whisper-large & 364711.2741 & 64.2591 & 15.3622 \\
        microsoft/Phi-3.5-mini-instruct & 0.5* & 2.0* & 0.5* \\
        microsoft/phi-1.5 & 12.9090 & 6.94e-10 & 9.6670 \\
        google/gemma-2-9b-it & 280939.5667 & 102.4015 & 25.2924 \\
        Qwen/Qwen2.5-7B-Instruct & 0.5* & 2.0* & 0.5* \\
        \bottomrule
    \end{tabular}
    }
    \caption{Summary of model parameters ($\lambda_i$, $\mu_i$, $\sigma_i$) for different top 50 models with the largest number of fine-tuned models. Here, ``*'' indicates the framework equation \ref{eq:citation_model} failed to fit the empirical data.}
    \label{tab:model_parameter_summary}
\end{table}

\section{Cumulative Trajectory of Finetuned Models By Organization} \label{sec:individual_cummulative_citation}

Understanding the cumulative fine-tuning trajectory of models across different companies provides valuable insights into the adoption dynamics and impact of open-source AI models. Fine-tuning is a critical mechanism through which base models are adapted to diverse applications, reflecting their versatility and the strategic priorities of the organizations that develop them. By examining the temporal evolution of fine-tuned models ($c^t$) for each company, we can identify influence patterns, competitive positioning, and the extent to which specific base models drive downstream innovations. The following analysis presents a comparative view of fine-tuning trends across companies, highlighting variations in growth rates, early adoption, and long-term sustainability in the AI ecosystem.
\begin{figure}[ht]
    \begin{center}
        \includegraphics[width=\linewidth]{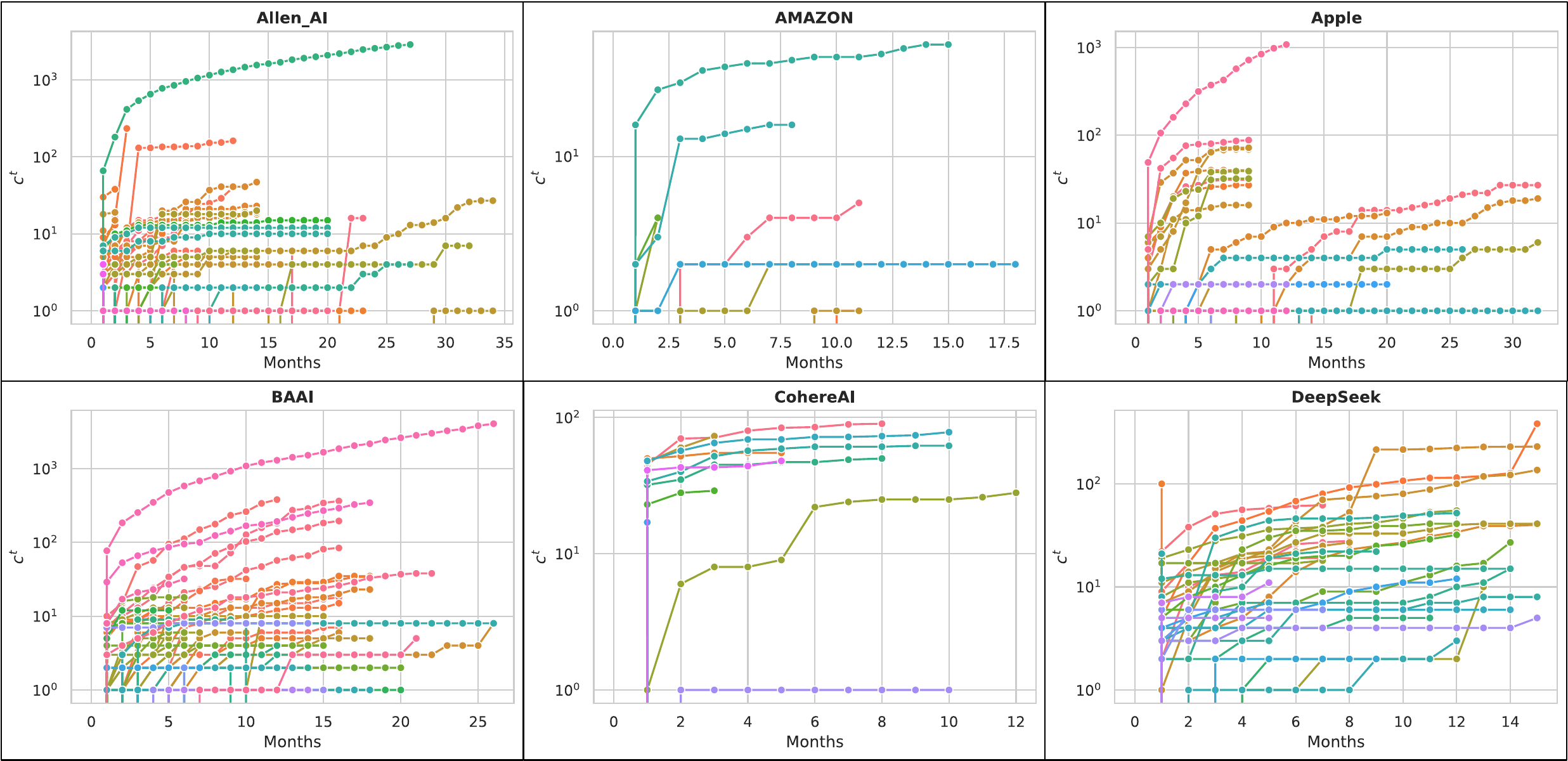}
    \end{center}
    \caption{The cumulative number of fine-tuned models (\( c_t \)) over time (months) for Allen AI, Amazon, Apple, Beijing Academy of Artificial Intelligence(BAAI), CohereAI and DeepSeek. }
    \label{fig:A_num_finetuned_company}
\end{figure}
\begin{figure}[ht]
    \begin{center}
        \includegraphics[width=\linewidth]{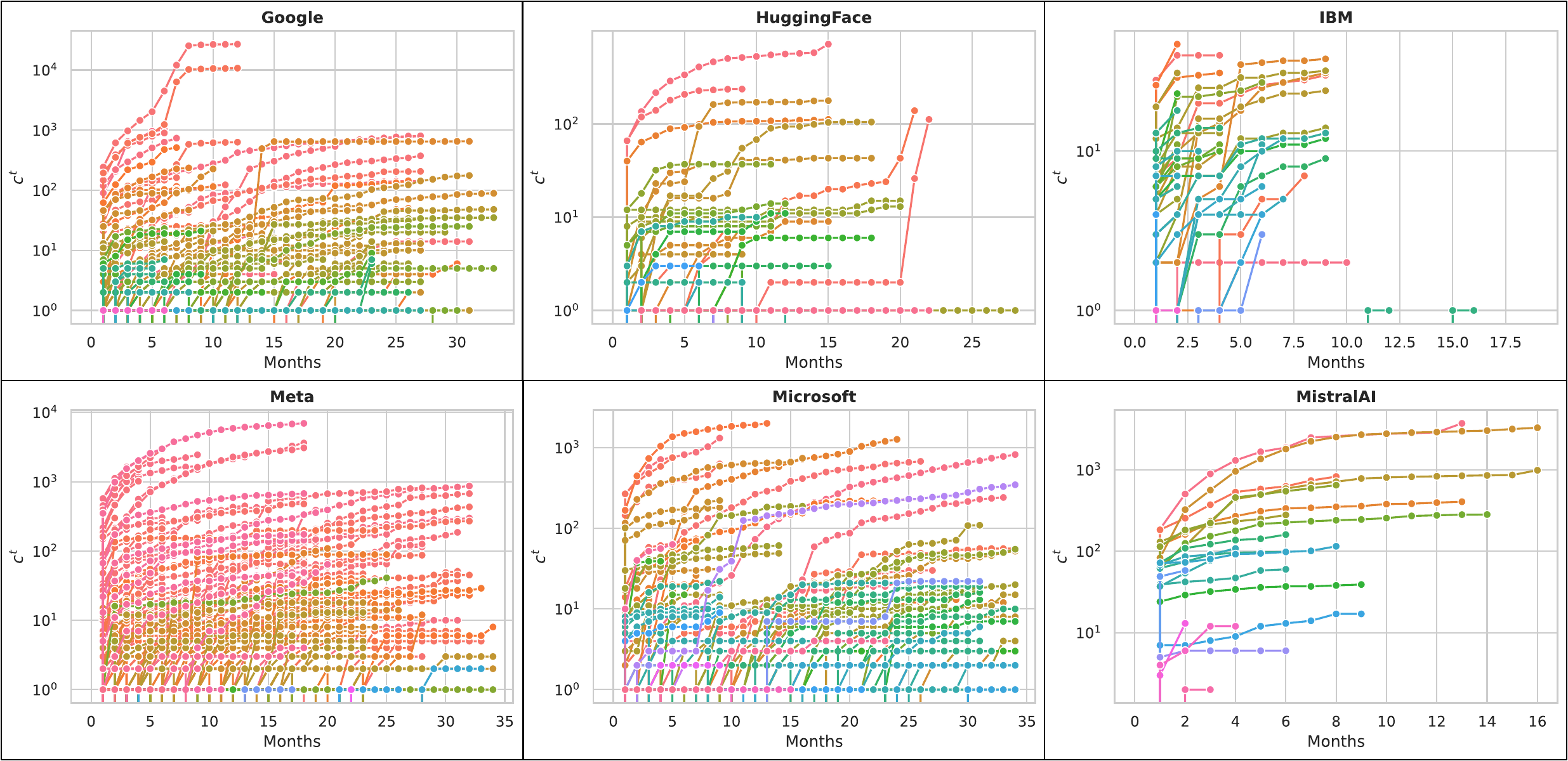}
    \end{center}
    \caption{The cumulative number of fine-tuned models (\( c_t \)) over time (months) for Meta, Google, HuggingFace, IBM, Microsoft, and MistralAI.}
    \label{fig:B_num_finetuned_company}
\end{figure}
\begin{figure}[ht]
    \begin{center}
        \includegraphics[width=\linewidth]{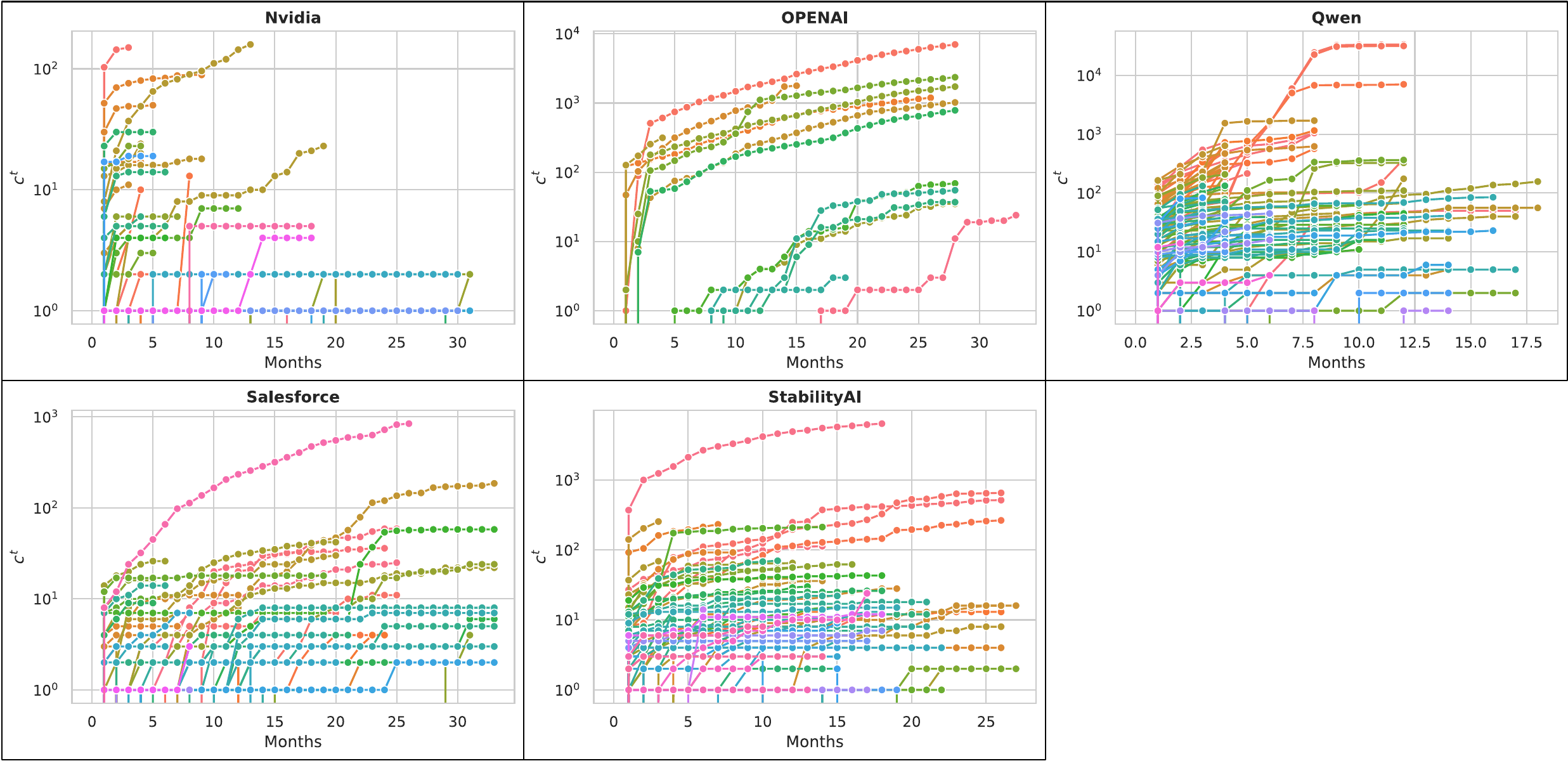}
    \end{center}
    \caption{The cumulative number of fine-tuned models (\( c_t \)) over time (months) for Nvidia, OpenAI, Qwen, Salesforce, and StabilityAI.}
    \label{fig:C_num_finetuned_company}
\end{figure}

\section{Analyzing the Cumulative Number of Downloads} \label{sec:downloads}
	
	 \begin{figure}[ht]
        		\begin{center}
          	 \includegraphics[width=\linewidth]{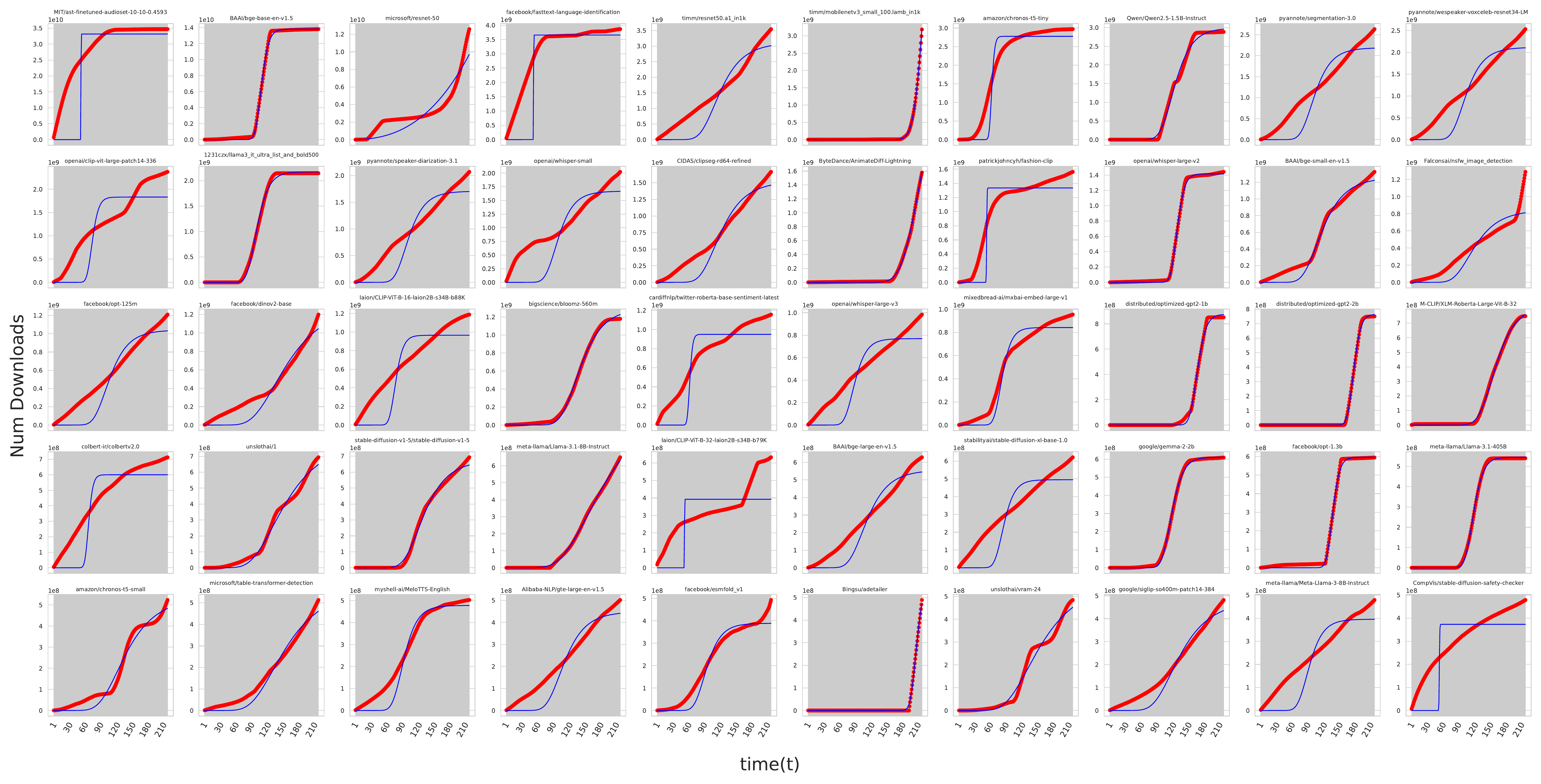}
        \end{center}
        \caption{The line plot of the cumulative number of downloads over time (day) for individual models ordered based on the most cumulative downloads. The blue plot is the predictive trajectory using the citation model.}
        \label{fig:cumulative_download_individual}
   	 \end{figure}
	
    In addition to looking at a number of fine-tuned models, we also briefly study the trajectory of the number of downloads for different models. As mentioned in Section \ref{sec:data_collection}, we only have download data of models after September 2024. Figure \ref{fig:cumulative_download_individual} shows that the framework utilized for downloads does not quite fit the framework well since the trajectories differ for different open-weight models. However, this is due to the framework being fitted on empirical data with an interval of data after the model's release; hence, it is fitted with incomplete data since the initial release day. 
        
\begin{figure}[ht]
        \begin{center}
            \includegraphics[width=\linewidth]{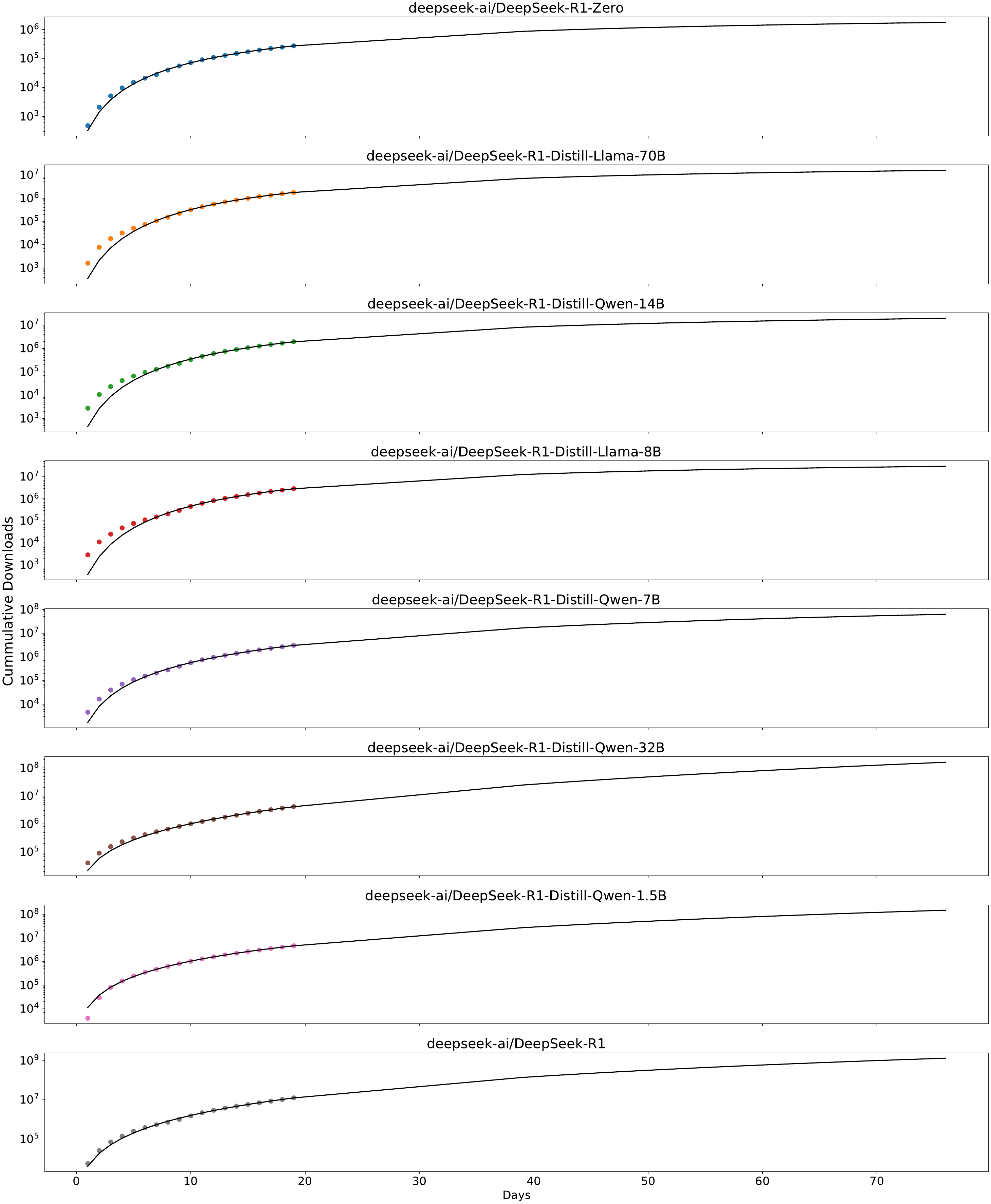}
        \end{center}
        \caption{Predicting number of downloads of recently popular DeepSeek models. The black line plot predicts the cumulative number of downloads of DeepSeek models up to 75 days after its release. }
        \label{fig:deepseek_downloads}

    \end{figure}

    Figure \ref{fig:deepseek_downloads} illustrates the prediction made for the cumulative downloads for the recently popular DeepSeek models. Colored markers in each subplot represent the observed download counts at different points in time, and the black curve superimposed on each set of points illustrates the model’s predicted growth trajectory. The vertical scale (cumulative downloads) spans several orders of magnitude, reflecting substantial variation in popularity among the model variants. Overall, the close alignment between the observed data and the predicted curves suggests that the modeling approach described in Appendix Section\ref{sec:framework} provides a reasonable fit for each DeepSeek variant’s download pattern. 

    From the figure, each model’s cumulative downloads follow the hypothesized saturating pattern over time, as depicted by the black curves. Early in the launch phase, downloads increase sharply, indicating rapid adoption. After several weeks, the model’s trajectory starts to flatten, showing a decrease in the download growth rate. We posit that as more advanced models become available, each DeepSeek variant’s download growth naturally slows off. Early adopters generate an initial spike, but user interest in earlier releases diminishes once new competitors emerge—potentially offering higher accuracy, additional features, or improved efficiency. This leads to the observed plateau in download rates, as existing users have already adopted the model, and new users may opt for more recent, better‐performing alternatives.

\begin{thebibliography}{10}

\bibitem{wangQuantifyingLongTermScientific2013}
Dashun Wang, Chaoming Song, and Albert-L{\'a}szl{\'o} Barab{\'a}si.
\newblock Quantifying {{Long-Term Scientific Impact}}.
\newblock {\em Science}, 342(6154):127--132, October 2013.

\bibitem{bommasaniOpportunitiesRisksFoundation2022}
Rishi Bommasani, Drew~A. Hudson, Ehsan Adeli, Russ Altman, Simran Arora, Sydney von Arx, Michael~S. Bernstein, Jeannette Bohg, Antoine Bosselut, Emma Brunskill, Erik Brynjolfsson, Shyamal Buch, Dallas Card, Rodrigo Castellon, Niladri Chatterji, Annie Chen, Kathleen Creel, Jared~Quincy Davis, Dora Demszky, Chris Donahue, Moussa Doumbouya, Esin Durmus, Stefano Ermon, John Etchemendy, Kawin Ethayarajh, Li~{Fei-Fei}, Chelsea Finn, Trevor Gale, Lauren Gillespie, Karan Goel, Noah Goodman, Shelby Grossman, Neel Guha, Tatsunori Hashimoto, Peter Henderson, John Hewitt, Daniel~E. Ho, Jenny Hong, Kyle Hsu, Jing Huang, Thomas Icard, Saahil Jain, Dan Jurafsky, Pratyusha Kalluri, Siddharth Karamcheti, Geoff Keeling, Fereshte Khani, Omar Khattab, Pang~Wei Koh, Mark Krass, Ranjay Krishna, Rohith Kuditipudi, Ananya Kumar, Faisal Ladhak, Mina Lee, Tony Lee, Jure Leskovec, Isabelle Levent, Xiang~Lisa Li, Xuechen Li, Tengyu Ma, Ali Malik, Christopher~D. Manning, Suvir Mirchandani, Eric Mitchell, Zanele Munyikwa, Suraj Nair,
  Avanika Narayan, Deepak Narayanan, Ben Newman, Allen Nie, Juan~Carlos Niebles, Hamed Nilforoshan, Julian Nyarko, Giray Ogut, Laurel Orr, Isabel Papadimitriou, Joon~Sung Park, Chris Piech, Eva Portelance, Christopher Potts, Aditi Raghunathan, Rob Reich, Hongyu Ren, Frieda Rong, Yusuf Roohani, Camilo Ruiz, Jack Ryan, Christopher R{\'e}, Dorsa Sadigh, Shiori Sagawa, Keshav Santhanam, Andy Shih, Krishnan Srinivasan, Alex Tamkin, Rohan Taori, Armin~W. Thomas, Florian Tram{\`e}r, Rose~E. Wang, William Wang, Bohan Wu, Jiajun Wu, Yuhuai Wu, Sang~Michael Xie, Michihiro Yasunaga, Jiaxuan You, Matei Zaharia, Michael Zhang, Tianyi Zhang, Xikun Zhang, Yuhui Zhang, Lucia Zheng, Kaitlyn Zhou, and Percy Liang.
\newblock On the {{Opportunities}} and {{Risks}} of {{Foundation Models}}, July 2022.

\bibitem{costaDemocratizationArtificialIntelligence2024}
Carlos~J. Costa, Manuela Aparicio, Sofia Aparicio, and Joao~Tiago Aparicio.
\newblock The {{Democratization}} of {{Artificial Intelligence}}: {{Theoretical Framework}}.
\newblock {\em Applied Sciences}, 14(18):8236, September 2024.

\bibitem{luitseGreatTransformerExamining2021}
Dieuwertje Luitse and Wiebke Denkena.
\newblock The great {{Transformer}}: {{Examining}} the role of large language models in the political economy of {{AI}}.
\newblock {\em Big Data \& Society}, September 2021.

\bibitem{euArtificialIntelligenceAct}
EU.
\newblock Artificial {{Intelligence Act}}, 2024.

\bibitem{maslejArtificialIntelligenceIndex2024}
Nestor Maslej, Loredana Fattorini, Raymond Perrault, Vanessa Parli, Anka Reuel, Erik Brynjolfsson, John Etchemendy, Katrina Ligett, Terah Lyons, James Manyika, Juan~Carlos Niebles, Yoav Shoham, Russell Wald, and Jack Clark.
\newblock Artificial {{Intelligence Index Report}} 2024, 2024.

\bibitem{hendersonFoundationModelsFair2023}
Peter Henderson, Xuechen Li, Dan Jurafsky, Tatsunori Hashimoto, Mark~A. Lemley, and Percy Liang.
\newblock Foundation {{Models}} and {{Fair Use}}, March 2023.

\bibitem{chanHazardsIncreasinglyAccessible2023}
Alan Chan, Ben Bucknall, Herbie Bradley, and David Krueger.
\newblock Hazards from {{Increasingly Accessible Fine-Tuning}} of {{Downloadable Foundation Models}}, December 2023.

\bibitem{chanVisibilityAIAgents2024}
Alan Chan, Carson Ezell, Max Kaufmann, Kevin Wei, Lewis Hammond, Herbie Bradley, Emma Bluemke, Nitarshan Rajkumar, David Krueger, Noam Kolt, Lennart Heim, and Markus Anderljung.
\newblock Visibility into {{AI Agents}}.
\newblock In {\em Proceedings of the 2024 {{ACM Conference}} on {{Fairness}}, {{Accountability}}, and {{Transparency}}}, {{FAccT}} '24, pages 958--973, New York, NY, USA, June 2024. Association for Computing Machinery.

\bibitem{eirasRisksOpportunitiesOpenSource2024}
Francisco Eiras, Aleksandar Petrov, Bertie Vidgen, Christian Schroeder, Fabio Pizzati, Katherine Elkins, Supratik Mukhopadhyay, Adel Bibi, Aaron Purewal, Csaba Botos, Fabro Steibel, Fazel Keshtkar, Fazl Barez, Genevieve Smith, Gianluca Guadagni, Jon Chun, Jordi Cabot, Joseph Imperial, Juan~Arturo Nolazco, Lori Landay, Matthew Jackson, Phillip H.~S. Torr, Trevor Darrell, Yong Lee, and Jakob Foerster.
\newblock Risks and {{Opportunities}} of {{Open-Source Generative AI}}, 2024.

\end{thebibliography}
\end{document}